\documentclass[journal]{IEEEtran}
\usepackage{color}
\usepackage{algorithm2e}
\usepackage{algorithmic}
\usepackage{pgfplots}
\usepackage{amsfonts}
\usepackage{amssymb}
\usepackage{mathtools}
\usepackage{bbm}
\usepackage{dsfont}
\usepackage{float, comment}

\ifCLASSINFOpdf
\else
\fi
\begin{document}
%
\title{Self-Segregating and Coordinated-Segregating Transformer for Focused Deep Multi-Modular Network for Visual Question Answering}
%
%
%

\author{Chiranjib~Sur \\ 
		Computer \& Information Science \& Engineering Department, University of Florida.\\
		Email: chiranjibsur@gmail.com
}

%
%

\markboth{Journal of XXXX,~Vol.~XX, No.~X, AXX~20XX}%
{Shell \MakeLowercase{\textit{et al.}}: Bare Demo of IEEEtran.cls for IEEE Journals}
%

\maketitle

\begin{abstract}
Attention mechanism has gained huge popularity due to its effectiveness in achieving high accuracy in different domains. But attention is opportunistic and is not justified by the content or usability of the content. Transformer like structure creates all/any possible attention(s). We define segregating strategies that can prioritize the contents for the applications for enhancement of performance. We defined two strategies: Self-Segregating Transformer (SST) and Coordinated-Segregating Transformer (CST) and used it to solve visual question answering application. 
Self-segregation strategy for attention contributes in better understanding and filtering the information that can be most helpful for answering the question and create diversity of visual-reasoning for attention. This work can easily be used in many other applications that involve repetition and multiple frames of features and would reduce the commonality of the attentions to a great extent. Visual Question Answering (VQA) requires understanding and coordination of both images and textual interpretations. Experiments demonstrate that segregation strategies for cascaded multi-head transformer attention outperforms many previous works and achieved considerable improvement for VQA-v2 dataset benchmark. 
\end{abstract}


%
\IEEEpeerreviewmaketitle

\section{Introduction} \label{section:introduction}

\begin{figure}[ht]
\vskip 0.2in
\begin{center}
\centerline{\includegraphics[width=.85\columnwidth]{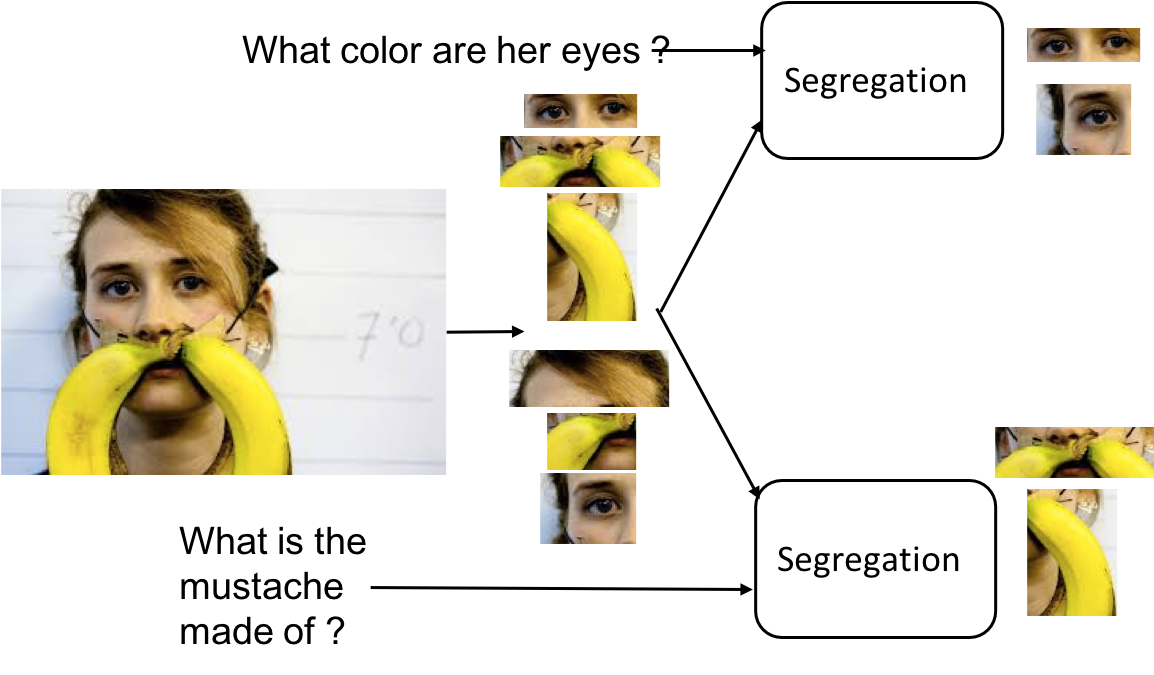}}
\caption{Explanation of Segregation.}
\label{fig:explain2}
\end{center}
\vskip -0.2in
\end{figure}

\begin{figure}[ht]
\vskip 0.2in
\begin{center}
\centerline{\includegraphics[width=.8\columnwidth]{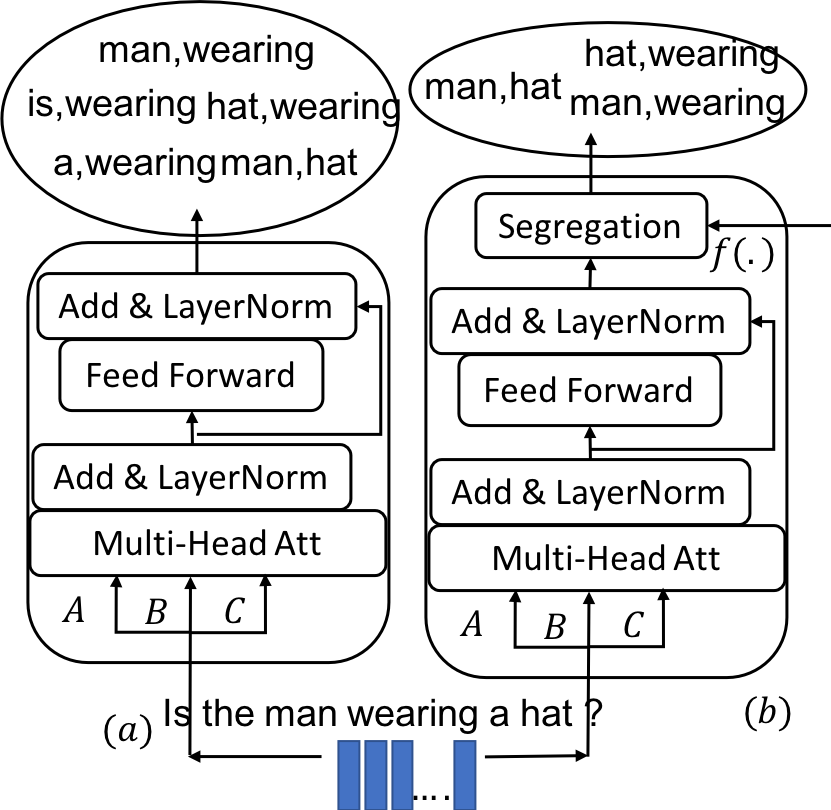}}
\caption{Explanation of Characteristics of Transformer as (a) and Segregation Transformer as (b).}
\label{fig:difference}
\end{center}
\vskip -0.2in
\end{figure}

\begin{figure}[ht]
\vskip 0.2in
\begin{center}
\centerline{\includegraphics[width=.95\columnwidth]{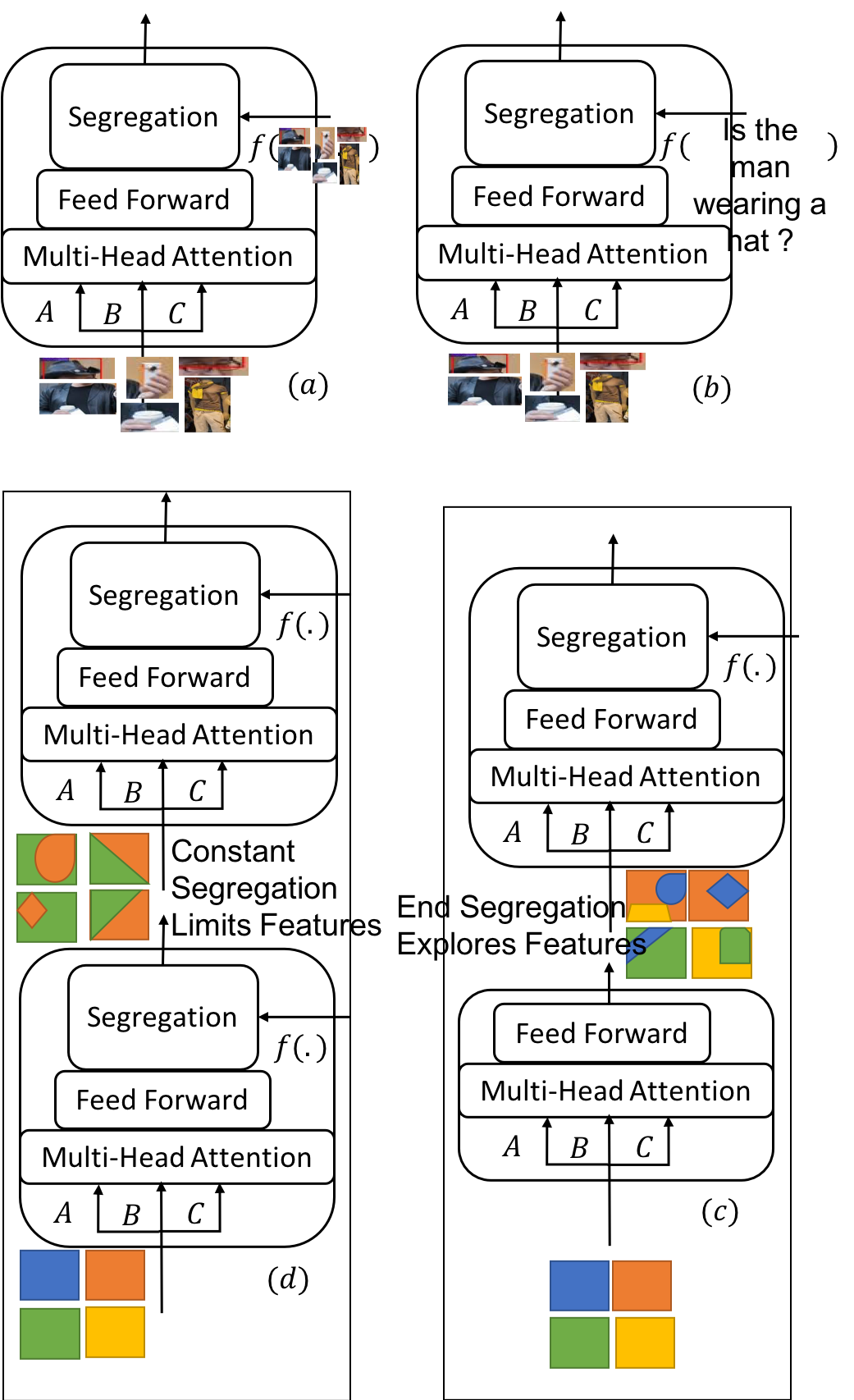}}
\caption{Explanation of Self-Segregating  (SST)  as (a) and Coordinated-Segregation  (CST) Approach as (b) End Segregation as (c) and Constant Segregation as (d). Segregation helps in propagation of important features, which can contribute more rational reasoning for VQA.}
\label{fig:explain}
\end{center}
\vskip -0.2in
\end{figure}

Visual Question Answering (VQA) bridges the gap between visual contents and natural languages through extrapolating the difference between the contents and reducing human involvement to deliver the query. 
Recent works in vision-language ranges from image captioning \cite{vinyals2016show}, 
image-text matching \cite{nam2017dual, kim2018bilinear}, visual captioning \cite{donahue2015long, xu2015show, anderson2018bottom}, visual grounding \cite{fukui2016multimodal, yu2018rethinking} and visual question answering (VQA) \cite{malinowski2014multi, kim2018bilinear, zhao2018open}.
A stacked attention network was proposed to learn different attention through iteration in  \cite{yang2016stacked}. 
Different multimodal bilinear pooling methods are proposed like in \cite{veit2016residual, fukui2016multimodal}. 
 \cite{anderson2018bottom} introduced a bottom up top down approach for images.
A co-attention learning framework to alternately learn the image attention and question attention
was introduced in \cite{lu2016hierarchical}. Some other works include multimodal fusion model to predict the answer \cite{zhou2015simple}, multi-stage coattention learning model to refine the attentions based on memory of previous attentions \cite{nam2017dual}, low-rank bilinear pooling for attention networks \cite{kim2016hadamard}, spatial memory network model for correlation estimation of every image patches and tokens in a sentence \cite{xu2016ask}, self-attention for a language embedding and question-conditioned attention for a visual embedding. A severe bottleneck is the lack of understanding of fine-grained relationships among multimodal features. Dense co-attention models can handle the bottleneck as in \cite{kim2018bilinear, nguyen2018improved} where the complete interaction is established between each question word and each image region. Our work exploited this correlation and used segregation on top of that.

VQA is a challenging task and is about inference of visual reasoning and understanding the complexity of the questions and its relation to visual representation. \cite{yu2019deep} provided a transformer based dual-attention model based on multi-head attention. Traditional transformer has limitations because of its limited ability to selection of features.  
Research in attention has progressed in vision \cite{nam2017dual}, language \cite{anderson2018bottom}, and attention models like dense co-attention models like \cite{kim2018bilinear}  and \cite{nguyen2018improved} have been proposed to model dense interactions between regional features of an image and question feature interface. Attention overcomes the problem of insufficient multi-modal interactions. Dense co-attention networks \cite{lu2016hierarchical},\cite{nam2017dual},\cite{xu2016ask} aids the understanding of image-question relationship for correct answers. Dense co-attention can be cascaded to create deeper representations and assists in potential improvements. 

\begin{figure*}
\vskip 0.2in
\begin{center}
\centerline{\includegraphics[width=1.85\columnwidth]{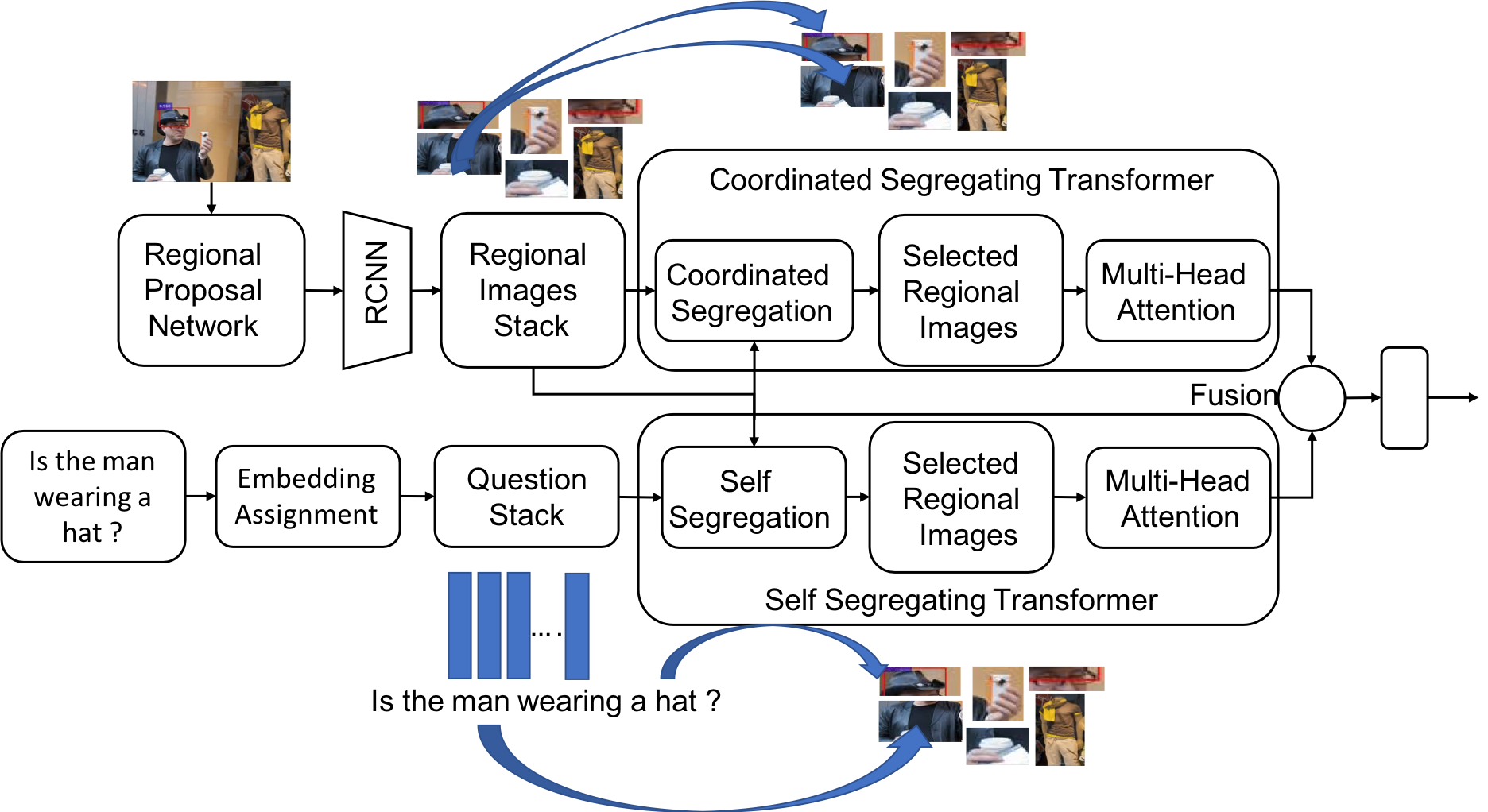}}
\caption{Architectural Details of Combination of Self-Segregating Transformer Model and Coordinated-Segregation Transformer Model for Visual Question Answering.}
\label{fig:an2}
\end{center}
\vskip -0.2in
\end{figure*}

\cite{yu2019deep} proposed a dense co-attention transformer model for attention using the regional features of \cite{anderson2018bottom}. This work reproduced the co-attention with transformer for better effectiveness, leveraging the ability to select the features that are much more relevant. 
Also, transformer model creates attention through combining all/any features, introducing lots of irrelevancy, inconsistency, and contamination. 
The irrelevancy is extracted either from the influence of image features or the natural language. What our novel transformer model introduced an extra query based on the content of the features and tries to identify them. It helps in composing another set of features that can identify the activities and their importance in an image. If we consider the feature space to be $ \{\textbf{v}_1, \ldots, \textbf{v}_n \}  \in \mathbb{S}$, then it is important to generate attentions to extract different combinations $ \left\lbrace \{ \textbf{v}_{x_{1}}, \ldots , \textbf{v}_{x_{m}}  \}_{}^{}, \ldots, \{ \textbf{v}_{y_{1}}, \ldots , \textbf{v}_{y_{2}}  \}_{}^{} \right\rbrace $ that can capture useful information from the attentions. However, models are biased to capture similar features $ \{ \textbf{v}_{x_{1}} , \ldots, \textbf{v}_{x_{m}}  \}_{}^{} \approx  \{ \textbf{v}_{y_{1}}, \ldots , \textbf{v}_{y_{2}} \} $ because of the trained weights. Present research considers encoding approaches \cite{yu2019deep} expecting that the important features will segregate. But it is a bad strategy  to encode through recurrent network function $\phi(.)$ as $\textbf{a}_{tt_{t}} = \phi(\textbf{v}_{t} , \textbf{a}_{tt_{t-1}})$ where the features overlap to generate the final features instead of intelligent combination. $\textbf{a}_{tt_{t}}$ is the encoded attention at time $t$. Encoding fails to scale and generalize well. Our approach creates an extra set of functionality that can help identify the attentions or features that are more helpful.

The fundamental problem, that we are trying to solve, is to counter the over-dependency of the models to combine multiple information. There are challenges that need to be solved to counter the lack of effectiveness when dealing with multiple frames/features. In VQA, there is the lack of identification of the feature(s) that can be helpful. Furthermore, several layers of non-linear approximation reduces the visibility of many attributes in the feature space of the objects. This creates misjudgment and biases in the network and is difficult to remove. In this work, we have proposed a self-segregation strategy that can help in better effectiveness and can reduce the model contents to a great extent. There are two kinds of strategies to identify what is useful. One is self-understanding and another is coordinated-understanding of the feature space. In visual question answering, we can identify these useful features, when attended them with more sincerity. This sincerity comes with self-segregation and we have identified this problem as very intriguing. This could reduce the feature space before the generation of the representation. This reduction in feature space can also help in better performance and reduce the burden of the weights to capture different attributes and intricacies of the image. Figure \ref{fig:explain} provide a comparative overview of different segregation techniques. 
Since dual attention \cite{nam2017dual} has been very successful in many applications for enhancing the performance, we have used both segregation strategies in such architecture to show its influence on inference.  Our architectures  have successfully outperformed many of the previous works or as good as them.  Later, we combined the segregated attention to generate the relevant tensors for the applications.

The rest of the document is arranged with 
architecture and details of implementation in Section \ref{section:architecture}, 
analysis of the experiments in Section \ref{section:results} and concluding remarks in Section \ref{section:conclusion}. 

The main contributions of this work are the followings. 1) strategies of segregation of the relevant information through self-segregation and coordinated-segregation in a deep dense co-attention Transformer network is introduced 2) solved the bottleneck of biases of multimodal interaction and any/all combinations of features selected for attention  3)  our segregation model helped in multi-modular co-attention model to perform better than many of the previous works 4) it is a simultaneous dense feature selector cum attention generation approach and has been introduced for the first time in the literature.


\section{Architecture} \label{section:architecture}
Transformer \cite{vaswani2017attention, lee2018set} has been successful in many applications including prediction and question answering \cite{devlin2018bert}, POS-Phrase prediction for sentences \cite{sur2020rbn}, due to its capability of multi-head attention generation through scaled dot-product. A few variants are introduced like Set transformer \cite{lee2018set}, evolved transformer \cite{so2019evolved}. This enhanced performance comes at a very high cost of parametric training and combination of all possibilities.  It considers different combinations of the features for preparation of the attentions, putting up different prospective. But transformer has some serious drawbacks, mainly, when the features are large scale. There is no way we can solve a problem through creating an infinitely large feature combination (visual-reasoning feature as well in VQA) space. We need to learn to identify and combine them effectively. For a feature set of dimension $d$, a multi-headed transformer network generates attention based on the adjacency of features in a set. In such scenario, there is no reason why the usable feature will segregate for all possibilities. Instead, there are chances that the relevant features get replaced with  highly approximated attention. This is evident from the nature of deep learning, where the weights fail to capture everything. Thus, creating attentions that are biased or combinations of wrong possibilities. This is where the representation does not generalize well. Our novel architecture and introduced concept help overcome through providing more scopes for generalization. We advocate a procedure that knows how to segregate relevant information through the use of the information itself and other means, like another relevant tensor. In VQA application, the question tensor is trainable and thus creates ample scope for re-structuring, while the image features are extracted through convolution network. We have introduced different strategies on how to implement segregation and discussed their limitations and advantages.  
Figure \ref{fig:difference} provided a pictorial overview of the novelty of the Self-Segregation Transformer, Coordinated-Segregation Transformer with the normal Transformer model. 

\subsection{Transformer Characteristics}
Transformer is about generation of attentions \cite{vaswani2017attention} through the combination of all/any possibilities.  
It provides very unique representations for applications related to natural language processing (NLP), but it fails to structure a visual composition because of the lack of tackling structural data for visual reasoning. Hence, we introduced strategies to extract relevant data and transfer them to posterior likelihood. Moreover, it is important for the network to recognize the relevant information instead of combining all of them blindly. In NLP, the adjacency of information holds immense information, which transformer captures. But, for image features, there are requirements of additional units to capture usable information, including attributes. 
Consider a transformer model with $n$ attentions of $d$ dimension, denoted by $f \in \mathbb{R}^{n \times d} $ generated from function $\psi(.) $ for $n$ different objects,  
\begin{equation}
 f = \psi(\textbf{A}, \textbf{B}, \textbf{C}) = \text{softmax} \left( \frac{\textbf{A}\textbf{B}}{\sqrt{d}} \right)\textbf{C}
\end{equation}
with $\textbf{A}   \in \mathbb{R}^{d \times d} $, $ \textbf{B} \in \mathbb{R}^{d \times d} $, $ \textbf{C} \in \mathbb{R}^{d \times d} $, we can recognize these matrices as combiner $ \textbf{A}$, combiner $ \textbf{B}$ to give rise to spatial-reasoning structural positioning space $ \textbf{A} \textbf{B}$. Spatial-reasoning structural positioning space captures the objects and map it to its corresponding attributes. $ \textbf{C}$ is enhancer or amplifier. Consider each row of $\textbf{A}$ to be orthogonal (in ideal scenario) to one column of $\textbf{B}$, only certain positions of $\textbf{A}\textbf{B}$  will get expressed and amplified through $\textbf{C}$. $\textbf{C}$ is also a weighted component for  $\textbf{A}\textbf{B}$.  We no longer can recognize this as query, key and value matrix, traditionally used for NLP applications. 
Considering a multi-head scenario $\psi_M(.)$, we define the mathematical equations as, 
\begin{equation}
 f = \psi_M(\textbf{A}, \textbf{B}, \textbf{C}) = [\text{head}_1, \text{head}_2, \ldots, \text{head}_h] \textbf{W}
\end{equation}
where we define $\text{head}_h$ as,
\begin{equation}
 \text{head}_h = \psi(\textbf{A}\textbf{W}_{A_h}, \textbf{B}\textbf{W}_{B_h}, \textbf{C}\textbf{W}_{C_h}) 
\end{equation}
But, we need to take care of the $ \textbf{A} \textbf{B}$ feature space as the aim is to make the model sensitive to any changes on $ \textbf{A} \textbf{B}$. Hence, we use another segregation strategy that can limit the expressing of unnecessary combinations.

\begin{figure*}
\vskip 0.2in
\begin{center}
\centerline{\includegraphics[width=1.85\columnwidth]{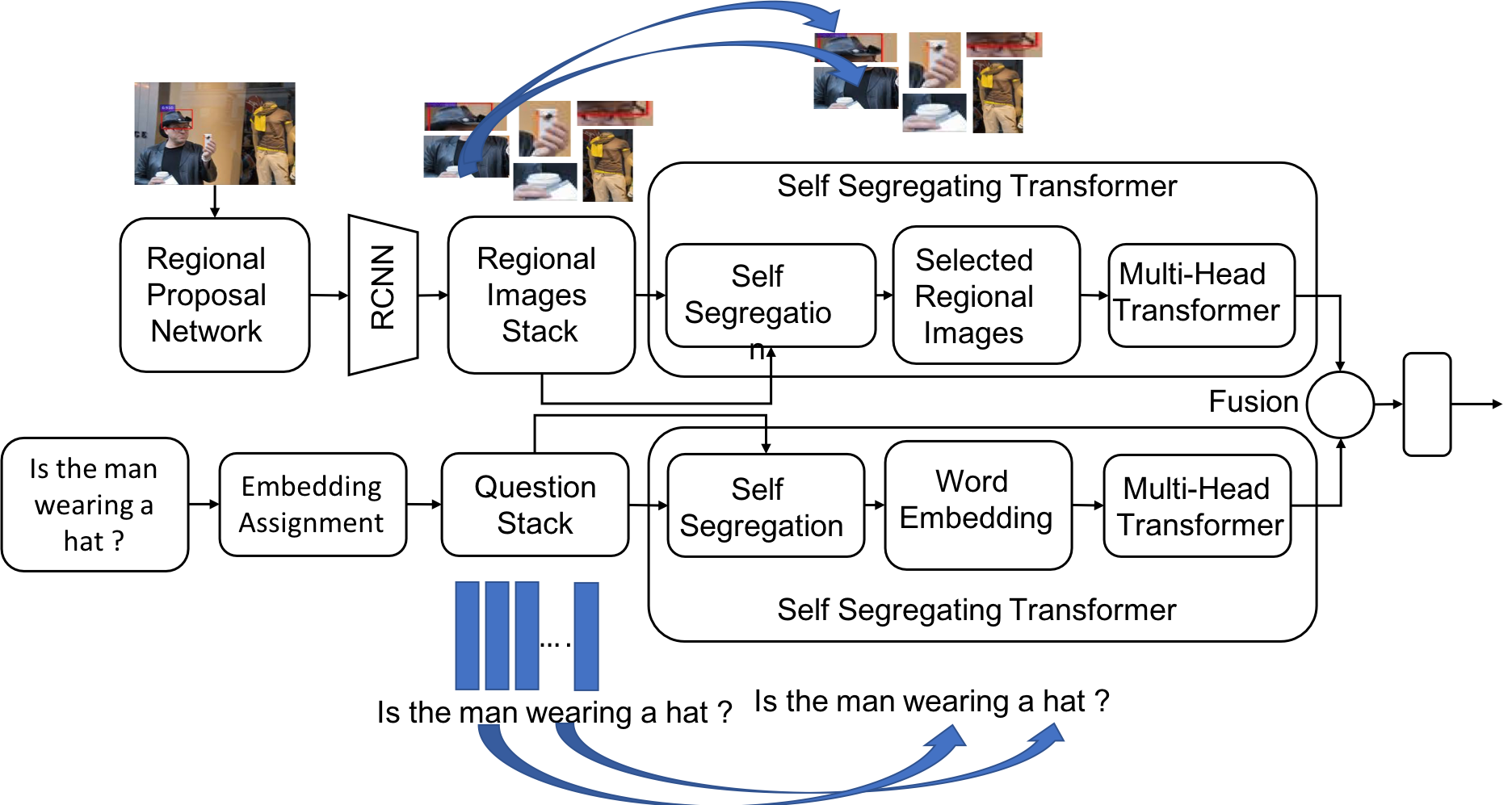}}
\caption{Architectural Details of Self-Segregating Transformer Model for Visual Question Answering.}
\label{fig:an1}
\end{center}
\vskip -0.2in
\end{figure*}

\subsection{Coordinated-Segregation Transformer} 
VQA requires visual-reasoning and is much more than attention combination.  We introduced segregation strategies to involve more reasoning and enhance the ability of the transformer to select a useful subset feature from the pool. Segregation prevents the limited capturing of information through linear transformation. Mathematically, for $\{ \textbf{q}_1, \ldots, \textbf{q}_m \}$ we can define it as the followings: we have $ \sum\limits_{i=1}^{m} \textbf{q}_i \textbf{W}_i  $ better than $ \sum\limits_{i=1}^{m} \textbf{q}_i a_i  $ with $\textbf{W}_i$ as linear transformer and $a_i$ as selectors.  In the latter, original space for $\{ \textbf{q}_1, \ldots, \textbf{q}_m \}$ is retained for semantic composition. In former, the attended compositions $\textbf{q}_i \textbf{W}_i$ approximates. 
Coordinated-Segregation Transformer (CST) is about relating the word embedding with the image features, where the two combiners are different. It promotes learning a better space for questions. At the same time, it is about training one combiner to get the tensor that can help in expressing the inference information in the $ \textbf{A} \textbf{B}$ feature space. $ \textbf{A} \textbf{B}$ defines the required visual-reasoning sub-space for expression of information. These expressions can be limited through Coordinated-Segregation function $\Phi(.)$. Limitation is important for learning representations with selectivity and specificity.  
The relevant features are made to express in the $ \textbf{A} \textbf{B}$ space and $\Phi(.)$ limits contamination, false negatives and false positives. 
Mathematically, the Coordinated-Segregation Transformer model can be defined as, 
\begin{equation} 
 f = \psi_M( \Phi(\omega, \textbf{A}, \textbf{B}), \textbf{B}, \textbf{C}) = [\text{head}_1, \ldots, \text{head}_h] \textbf{W}
\end{equation}
where we defined $\text{head}_h$ as,
\begin{equation}
 \text{head}_h = \psi( \Phi(\omega, \textbf{A}, \textbf{B})\textbf{W}_{A_h}, \textbf{B}\textbf{W}_{B_h}, \textbf{C}\textbf{W}_{C_h}) 
\end{equation}
We have defined segregation function $\Phi(.)$  as the following, 
\begin{equation}
\begin{split}
  \Phi(\omega, \textbf{A}, \textbf{B}) & = [ \Phi(\omega_1, \textbf{A}, \textbf{B}) \text{ } ; \text{ }  \ldots \text{ } ; \text{ } \Phi(\omega_h, \textbf{A}, \textbf{B})  ]  \\
   & =  [ \omega_1 * \textbf{A} \text{ } ; \text{ } \ldots \text{ } ; \text{ } \omega_h * \textbf{A} ]
  \end{split}
\end{equation}
where we defined segregator coefficients $\omega \in \mathbb{R}^{d}_{0/1}$ as,  
\begin{equation} \label{eq:omega}
 \omega_i = \sigma ( \textbf{U}_{_{AB_i}} \rho( \textbf{U}_{_{A_i}} \textbf{A} + \textbf{U}_{_{B_i}}\textbf{B}) )
\end{equation}
Segregator coefficients $\omega_i$ is generated from the tensors $\textbf{A}$ and $\textbf{B}$ and the weights $\textbf{U}_{_{AB_i}}$, $\textbf{U}_{_{A_i}}$  and $\textbf{U}_{_{B_i}}$. Equation \ref{eq:omega} establishes the understanding of the image data contents with the question feature space (word embedding here). 
We have $\sigma (.)$ and $\rho (.)$ as two functional approximations. $ \omega $ is expected to be a sparse matrix with high weight-age on the useful feature spaces. $ \omega $ is open to other kinds of functions that segregate features and enhance the performance. $\textbf{U}_{_{AB}}$ helps in generating the weight-age factor, $\textbf{U}_{_{A}}$ aids extracting information from the visual-reasoning features  and $\textbf{U}_{_{B}}$  generates the relevant counterparts from another learned feature space like NLP. Figure \ref{fig:an2} provided a pictorial overview of the Coordinated-Segregation Transformer architecture where Coordinated-Segregation is used. Apart from learning to extract useful frames, Equation \ref{eq:omega} establishes the difference between the image and question and make it useful for determination of the useful frames. If we consider repetitions of features in image frames, the final reasoning semantic tensor will have over-expressed points, while, in case of no-repetition, lightly expressed regions are amplified with $\textbf{C}$. The joint equation for the CST model, used for experiments, can be denoted as $f_{_{_{CST}}}$ and is defined as the followings.
\begin{equation} \label{eq:arch1}
\begin{split}
 f_{_{_{CST}}} = & Q_L(w \mid \{h_1,\ldots,h_k\}, \{g_1,\ldots,g_k\} )  \\
 &  Q_{A_1}( \{h_1,\ldots,h_k\} \mid \{q_1,\ldots,q_{M}\}) \\
 &  Q_{A_2}(\{g_1,\ldots,g_k\} \mid \{q_1,\ldots,q_{M}\} \\
 & ,  \{I_1,\ldots,I_N\})  Q_{_{RCNN}}(I_1,\ldots,I_N \mid \textbf{I})  
\end{split}
\end{equation}
where we have $Q_L(.)$ as the language decoder, $Q_{A_1}(.)$, $Q_{A_2}(.)$ are the segregation transformers, $Q_{_{RCNN}}(.)$ is the RCNN feature extractor. Also, $\{h_1, \ldots, h_k\}$, $\{g_1, \ldots, g_k\}$ are the multi-head attentions, $\{w\}$ is the generated answers, $\{I_1, \ldots, I_k\}$ are the regional image features for image $\textbf{I}$ and $\{q_1, \ldots, q_k\}$ are the question based language features.

\subsection{Self-Segregating Transformer}
Self-Segregation Transformer (SST) is an added advantage over self-attention \cite{vaswani2017attention} for reasoning semantic tensor generation. In SST, the coordinated term is replaced with self-information $ \textbf{A} \approx \textbf{B}$ in Equation \ref{eq:omega}. Figure \ref{fig:explain} provided a diagrammatic explanation of the SST. Self-Segregation is about identification of the relevant information before using the information for self-attention. Like any system, that is related to identifying the feature space, Self-Segregation is dependent on self-discovery of the relevant information that can help. It is unlike Coordinated-Segregation, which helps in creating influence. As we provide a fine-tuning strategy for NLP, the word embedding gets shape that it can relate to visual structures for the application. Figure \ref{fig:an2} and Figure \ref{fig:an1} provided diagrams of dual attention architecture \cite{yu2019deep} where Self-Segregation is used and experimental details and analysis of results are provided in Section \ref{section:results}. 
Another important utility of segregation is the constant object feature disappearance (important features propagate and unimportant features blocked) that can act as regularization for training these kinds of networks. This increases the model sensitivity and the model adapts dealing with lower regional features. The inference deals well with the segregated image features. 
The joint equation for the CST model, for the experiment, can be denoted as $f_{_{_{SST}}}$ and is defined as the followings.
\begin{equation} \label{eq:arch2}
 \begin{split}
  f_{_{_{SST}}} = & Q_L(w \mid \{h_1,\ldots,h_k\}, \{g_1,\ldots,g_k\} )  \\
 &  Q_{A_1}( \{h_1,\ldots,h_k\} \mid \{q_1,\ldots,q_{M}\}) \\
 &  Q_{A_2}(\{g_1,\ldots,g_k\} \mid \{I_1,\ldots,I_N\})  \\
 &  Q_{_{RCNN}}(I_1,\ldots,I_N \mid \textbf{I})  
\end{split}
\end{equation}
Notations defined in Equation \ref{eq:arch1}.

\subsection{Constant Segregating Transformer and End Segregation Transformer Framework}
Segregation can help in the determination of the usefulness of features that can be allowed to be expressed in the attention tensors provided it enhances the inference accuracy.  This segregation can be allowed to go on after each transformer layer or can be restricted only at the end. This gives rise to the strategies for Constant Segregation Transformer (CS) and End Segregation Transformer. Mathematically, we can denote Constant Segregation $\Phi_{_{CSeT}}(.)$ and End Segregation $\Phi_{_{ESeT}}(.)$ as the following equations. 
\begin{equation}
 \Phi_{_{CSeT}}(.) =  \Phi( \psi_{M_m}(\ \ldots \Phi( \psi_{M_1}(\textbf{A}, \textbf{B}, \textbf{C})) \ldots, \textbf{B}, \textbf{C}) ))
\end{equation}
\begin{equation}
 \Phi_{_{ESeT}}(.) = \Phi(   \psi_{M_m}( \ldots \psi_{M_1}(\textbf{A}, \textbf{B}, \textbf{C})) \ldots, \textbf{B}, \textbf{C}) ))
\end{equation}
for $m$ layers of transformer. 
The segregation concept can have different implications on the model on how it segregates the features for improvement. With that, we have introduced two different schemes that defined different layered multi-head transformer models. Segregation, in Constant Segregation Transformer Framework, occurs at each level for better capture and establishment of the structural integrity of the features. It is much focused segregation with initial decision and no external influence. There are chances of suppression of individuals at the beginning with limited compositional characteristics. Constant Segregation is not prone to excessive content and best for generative tasks like sequence generation, language generation. 
On the contrary, for exclusive search based strategies, End Segregation Transformer Framework is better. End Segregation is much more search based, exploratory with reasoning, decision at the end. External influences (like questions) will help in more explorations, with limited suppression of individual at the beginning. End Segregationcan have representation characteristics, can be prone to excessive contents and best for large-scale classification tasks like VQA, Question Answering, etc.



\subsection{Feature Composition Decoder}
Feature Composition Decoder operates on the attention generated from the segregation transformer model. For that, we have used different kinds of decoder schemes for each of the $n$ attentions $\{A_1,\ldots,A_n\}$. The first one is encoder version $E_t$ like the following equation, 
\begin{equation}
 F = E_n \text{ for } E_t = \Upsilon(A_{t-1},\textbf{W})
\end{equation}
where $F$ is the encoded tensors. Depending on applications, the dimension of $F$ varies. The other version is the multi-layer transformation version like the followings,  
\begin{equation}
  F = \sum\limits_{i}^{} \alpha_i A_i
\end{equation}
where the set $\alpha = \{\alpha_1,\ldots, \alpha_n\}$ with $\sum \limits_{i}^{} \alpha_i = 1$. With $\text{MLP}(.)$ as multi-layer perceptron function and $ \text{softmax}(.)$ as softmax layer, we have defined $\alpha$,
\begin{equation}
 \alpha = \text{softmax} (\text{MLP}([E_1,\ldots,E_n]))
\end{equation}
 The final output is extracted as the followings,
\begin{equation}
 \textbf{z} = (\beta_1 F_1\textbf{W}_1 + \beta_2F_2\textbf{W}_2) \textbf{W}_3 = \textbf{W}_{k+1} \sum\limits_{i=1}^{k} \beta_i \textbf{W}_{i} F_i
\end{equation}
where $F_1$ and $F_2$ are part of the multiple-attention framework, with $\sum \beta_i = 1$ for normalized feature representation.  \cite{yu2019deep} has provided the argument that encoding is much better in performance, since it aids in generating the best possible structural overview of the representation tensor for the answers. For experiments, we have trained the coefficients for maximization of the likelihood for answers and outperform existing works.

\section{Experiments, Results \& Discussion} \label{section:results}

\subsection{Dataset Description}
Experiments were conducted on the largest VQA benchmark dataset, VQA-v2 to evaluate the performance of our novel models.  It is based on MSCOCO images and contains human-annotated question-answer pairs relating to the images, with 3 questions per image and 10 answers per question. The dataset is split into training (80k images and 444k QA pairs), validation (40k images and 214k QA pairs) testing (80k images and 448k QA pairs). The results can be further micro-visualized with sub-division of 3 further types, namely Yes/No, Number, and Other with an overall accuracy. 

\subsection{Model Discussion}
For our experiments we have adopted a self/coordinated segregating multi-modular multi-layer co-attention model consisting of self-attention and guided-attention for generating the fine-grained features space for VQA.  This is based on the scaled dot-product attention based work \cite{vaswani2017attention}. 
Though transformer is related to queried attentions, we interpret the attention model as maximization of the related $\textbf{A}\textbf{B}$ feature sub-space for visual reasoning. It  establishes the relational coherence of the image features and learn-able natural languages. Here, the tensors with minimum cosine distance will establish very high amplitude and a softmax reduces the influences of any other than maximization. The tensor is divided by $\sqrt{d}$ (dimension of query) and prevents the overexposure of the closeness of the distances between image features and languages. 
Though the original transformer is proposed to have a query of  dimension $d$,  query can be assumed to be $\{ q_1 ,\ldots, q_n \}$, though in realty  self-attention concept \cite{vaswani2017attention, yu2019deep} has no physical meaning and utilization is limited. Our approach of segregation of information is much more justified and realistic. For state-of-the-art performance, it is always better to adopt a multi-modular co-attention model where self-attention $A_1$ is the relationship between the different objects present in the image. More chances of a relationship brings two segments very close to each other, like "white-trousers" image and "trousers" image will be close to each-other and their inner product will be maximum. Similarly for guided-attention, it is assumed that the inner product of tensors for image of  "white-trousers"  will be same for word embedding of "trousers", but it is very hard to control such operations. It can excite other false establishments. But our segregation approach will create a check on that through learning the feature space and emphasizing on usability and relevancy. 
Attention creates representation that converges for both regional image set and the word embeddings of the texts. However, in absence of concrete usability formulation (like segregation), multi-head attention creates several possibility and then encodes to the final tensor. These approaches are either encoding of events or stacking of events. Both the features are combinations of all possibilities.  

In our experiments, we have used the mean-pooled convolutional features of detected regions and is obtained from a ResNet architecture as image features, having dimension 2048.  GloVe word embedding of dimension 300 are used for initialization of embedding and regional image features of size $n \in [10, 100]$ \cite{anderson2018bottom} are used. Question sentences are of size $m \in [1,14]$ with dimension 300 pre-trained GloVe embedding. We have used masks for nullifying the effects of zero-padding for both question sentences and regional image embeddings. 
From the results in Table \ref{table:an1}, we can claim that our novel contribution helped in providing better representation and visual-reasoning tensors for VQA application.  The fundamental problem, that it solved, is the way the multiple features are dealt with for establishing hand-in-hand resemblance of visual-reasoning and language. While previous works considered weighted transformation for multi-frame features, we argue that segregation will refine the individuals and block the unnecessary. Segregated attentions have better effects on the performance than any other previous model performance. The segregation strategy gets the image adapted to a reasoning phase without contamination and focused on the questions representation for maximization of the likelihood.

\subsection{Evaluation Metrics}
VQA analysis is most dependent on evaluation metrics like misclassification error on different categorical groups like Y/N, Number and Other.  The sub-categorical information of the different samples in testing dataset can provide better overview of the performance for different question formulations. The sub-category performance promotes more refinement of the performance of the models at a finer level. Specific categorical improvements can be recorded from such analysis. It will point out the limitations of the models and can be better judged for further improvements.

\subsection{Implementation Details}
The hyper-parameters of the network are important for training. We have considered the latent dimension of the multi-headed transformer attention to be $ d = 512 $, while image dimension is 2048, question dimension with $300$ word embedding and final likelihood attention as 1024. We have considered the number of heads as $h=8$ and the answer vocabulary as $N= 3129$ with the depth of the Segregation Transformer to be $8$. For applications like VQA, where visual-reasoning is important, $h=8$ is reasonable with 6 layers of SST/CST for a better reasoning sub-space representation. To train the model with used  Adam optimizer with value of momentum as $ \beta_1 = 0.9$ and $\beta_2 = 0.98$ and learning rate to be $0.0001$ with decay rate of $1/2$ after every epoch. The batch size considered as 33/64 and 20 epoch is considered for analysis. 
Other important factors that were considered for the experiments were the dimension of the attention and the length of the question. The attention was kept at 512 for convergence in a reasonable time and compare with existing works. Existing works like \cite{vaswani2017attention} reported no improvement for 1024 dimension due to the saturation of the feature space without segregation. But, we found that the model improved with the increase in attention dimension. 

Encoder based features are used with word embedding for each word of the question. If we consider the words as $\{w_1,\ldots,w_{14}\}$, the word embedding representation is denoted as $[w_1\textbf{W}_E;w_2\textbf{W}_E;\ldots;w_{14}\textbf{W}_E]$, while the encoder based representation is denoted as $[E_n(w_1\textbf{W}_E, \textbf{0});E_n(w_2\textbf{W}_E, E_n(w_1\textbf{W}_E, \textbf{0}));\ldots;E_n(w_{14}\textbf{W}_E,$ $ \ldots)]$ for the $\textbf{W}_E$ word embedding set and encoder function $E_n(.)$. 
Our architectures have witnessed 2\% increase in accuracy when encoding based representation of the sentences are used instead of the original word embedding. The pre-trained word embedding has enhanced the performance than the initialized word embedding and fine-tuned with training of the network.  We have trained with only VQA Training Set to be evaluated with VQA Validation set and Test-dev set in the server. 


\begin{table}[t]
\caption{A Comparative Study of Different Schemes Introduced and Experimented with Self-Segregation and Coordinated-Segregation Transformer Architecture for Validation Set. Trained with only VQA Training Set.}
\label{table:an1}
\vskip 0.15in
\begin{center}
\begin{small}
\begin{sc}
\begin{tabular}{|l|c| c| c| c| c|}
\hline
Model-Framework  & All & Y/N & Num & Other    \\\hline 
CST+$\overline{\text{Image}}$+ESeT & 61.56  & 78.11 &  41.21  & 52.30  \\
CST+$\overline{\text{Image}}$+CSeT &  61.91  & 79.43 &  41.41  & 52.51   \\
SST+$e({\text{Image}})$+ESeT & 63.28 & 81.35 & 45.31  &  55.72  \\
SST+$e({\text{Image}})$+CSeT & 64.23 & 82.12 & 46.19 &  56.32   \\
SST+${\text{Ques}}$+ESeT & 63.09 & 81.29 & 45.31 &  55.61   \\
SST+${\text{Ques}}$+CSeT & 64.56 & 82.61 & 46.25 &  56.22   \\
SST+$\overline{\text{Image}}$+ESeT & \textbf{63.90} & \textbf{81.56} &  \textbf{46.62}  &  \textbf{54.44}  \\
SST+$\overline{\text{Image}}$+CSeT & \textbf{64.78}& \textbf{82.89} & \textbf{46.54} & \textbf{55.29}  \\\hline
\end{tabular}
\end{sc}
\end{small}
\end{center}
\vskip -0.1in
\end{table}

\begin{table*}[h]
\caption{A Comparative Study of the Best Self-Segregation and Coordinated-Segregation Transformer Architecture with Existing Works for Validation Set. Trained with only VQA Training Set.}
\label{table:an2}
\vskip 0.15in
\begin{center}
\begin{small}
\begin{sc}
\begin{tabular}{|l|c| c| c| c| c|}
\hline
Model-Framework  & All & Y/N & Num & Other    \\ 
\hline\hline
Up-Down \cite{anderson2018bottom} & 63.2 & 80.3 & 42.8 & 55.8   \\
MCAN$^*$ \cite{yu2019deep} & 62.86 & 80.51 & 45.91 & 53.92   \\
\hline
Ours$^*$ & \textbf{64.78}& \textbf{82.89} & \textbf{46.54} & \textbf{55.29}  \\\hline
\multicolumn{5}{l}{*Compared on Similar Model and Same Code Base.}\\
\multicolumn{5}{l}{**Trained on VQA (Training + Validation) and Visual Genome Training Data.}
\end{tabular}
\end{sc}
\end{small}
\end{center}
\vskip -0.1in
\end{table*}

\begin{table*}[h]
\caption{A Comparative Study of the Best Self-Segregation and Coordinated-Segregation Transformer Architecture with Existing Works  for Test-dev Set. Trained with only VQA Training Set.}
\label{table:an3}
\vskip 0.15in
\begin{center}
\begin{small}
\begin{sc}
\begin{tabular}{|l|c| c| c| c| c|}
\hline
Model-Framework  & All  & Other & Y/N   & Num  \\ 
\hline\hline
Up-Down** \cite{anderson2018bottom} & 65.32 &  56.05 & 81.82   & 44.21  \\
 MFH** \cite{Yu2017BeyondBG} & 68.76  &  54.84 & 84.27 &  49.56   \\
 MCAN$^*$ \cite{yu2019deep} & 65.51  & 55.16 & 81.69 & 49.41    \\
\hline
Ours$^*$ & 66.72  & 56.12  & 82.16  &  49. 77 \\\hline
\multicolumn{5}{l}{*Compared on Similar Model and Same Code Base.} \\
\multicolumn{5}{l}{**Trained on VQA (Training + Validation) and Visual Genome Training Data.}
\end{tabular}
\end{sc}
\end{small}
\end{center}
\vskip -0.1in
\end{table*}

\subsection{Result Analysis}
Experiments are defined on a finer scale to capture the influence of our novel architecture for VQA application. VQA itself requires a very fine-grained features that can capture complex relational and reasoning features. 
We have defined a few types of Segregation Variants for your experiments to figure out the influence of segregation at different levels. In Table \ref{table:an1}, we have provided the results for different segregation techniques. The same multiple-attention model is used for each of the Model-Framework. One series of attention is the language embedding $\textbf{L}$ and another is regional image features $\textbf{R}$. In CST, both the $\textbf{L}$ and $\textbf{R}$ are self-segregated with ESeT and CSeT. But in SST, $\textbf{L}$ is self-segregated, while $\textbf{R}$ is segregated with either the encoded image feature $enc({\text{Image}})$ or with language information $\textbf{L}$ or the mean of regional features $\textbf{R}$ as $(\overline{\text{Image}})$. We have $\frac{1}{N}\sum\limits_{N}^{i=1}  R_i $ $\forall $ $\{ R_1, \ldots R_N \} \in \textbf{R}$ and is denoted as $\overline{\text{Image}}$ in Table \ref{table:an1}.  
While, ${\text{Image}}$ and $\overline{\text{Image}}$ represents the overall image in some defined space, fine-trained language information $\textbf{L}$ must converge to fixed regional image sub-space. Compared to $enc({\text{Image}})$, $\overline{\text{Image}}$ performed much better as they weighted average scheme had a better feature spaced defined with alignment with the individual features $R_i$. Also, when it comes to defining the natural language space $\textbf{L}$ for ${\text{Ques}}$, it is subjective to the gradient and there is no individual provided. In fact, when converting the $300$ dimension to the required $512/1024$ dimension of the language, there is requirement of specific learning for improvement. This can be a future work strategy for more refinement of the model. CSeT has proven to be better than ESeT and has been demonstrated in our work. However, the initial transformer should allow all the features to propagate before imposing the segregation strategy. Another reason, why ESeT is as better as CSeT is because of the visual-reasoning requirement of VQA, which is hard to achieve for applications. Later, we have compared our best model with the previous works in Table \ref{table:an2}. Our work is coded on top of the code base of \cite{yu2019deep} and have reported the experimented baseline and the improvement. From Table \ref{table:an2}, we can say that there are scope of finer improvements when we introduce the segregation strategy in the mid of any multi-frame attention block of a network. 
Table \ref{table:an1} provided a comparative study of different segregation architectures only on the validation set and trained with only training set of the VQA dataset. 
From Table \ref{table:an1}, it can be easily concluded that ESeT is much better than CSeT because of the fact that it explores through the attention space for better reasoning, in comparison to others. The frequent segregation of the features can have better impact on the inference. Segregation scheme is also better than the encoding based visual-reasoning. Encoding based visual reasoning suffers from lack of proper representation and adjacency of features in the feature stream.
In Table \ref{table:an3}, we have provided results related to the Test-dev Set in the server and is better than the considered base model, to provide us with evidence that our approach can provide improvements, when added as  components in any network involving multiple frames.




\section{Conclusion} \label{section:conclusion}
This work introduced a novel transformer architecture called Self-Segregation and Coordinated-Segregation Transformer for VQA application. It is capable of segregating the usable visual-reasoning information for attention.  It consists of layers of segregation transformers and fusion of segregated tensors for generation of visual-reasoning attentions. This filtered attention scheme, when estimated, contributes to enhanced performance and is one of the state-of-the-art architecture for VQA application with more control of the language-image intermediate feature sub-space. It is a strategy to identify the usefulness and importance than process any/all. Processing thousands of features are expensive and deep neural networks do not guarantee extraction of the best. Hence, segregation of the usable tensors is an important strategy for many applications. This strategy will help in reducing the feature space for better representation learning and less contamination from the unnecessary ones. Learning to segregate useful information assists in outstanding identification of the attributes, related visual structures and infer visual-reasoning space for better answering of visual questions.

\ifCLASSOPTIONcaptionsoff
  \newpage
\fi

%








\end{document}